%% file: paper2035.tex
\documentclass[runningheads]{llncs}
\usepackage{graphicx}
\usepackage{bm}
\usepackage{amsmath}
\usepackage{color,soul}
\usepackage{hyperref}
\usepackage[misc]{ifsym}

\makeatletter
\newcommand{\smalloplus}{\mathbin{\mathpalette\make@small\oplus}}
\newcommand{\smallotimes}{\mathbin{\mathpalette\make@small\otimes}}

\newcommand{\make@small}[2]{%
  \vcenter{\hbox{%
    \scalebox{0.8}{$\m@th#1#2$}%
  }}%
}
\makeatother
\begin{document}
\title{3D Teeth Reconstruction from Panoramic Radiographs using Neural Implicit Functions}
\titlerunning{Occudent: 3D Teeth Reconstruction from Panoramic Radiographs}
%
\author{
Sihwa Park\inst{1} \and
Seongjun Kim\inst{1} \and
In-Seok Song\inst{2}  \and
Seung Jun Baek\inst{1(\textrm{\Letter})}
}
\authorrunning{S. Park et al.}
%
\institute{
Korea University, Seoul, South Korea\\
\email{\{sihwapark, iamsjune, sjbaek\}@korea.ac.kr}\\
\and
Korea University Anam Hospital, Seoul, South Korea\\
\email{densis@korea.ac.kr}\\}
\maketitle              
\renewcommand{\thefootnote}{}
\footnotetext{\inst{\textrm{\Letter}}Corresponding Author}
\renewcommand{\thefootnote}{\arabic{footnote}}
\setcounter{footnote}{0}
\begin{abstract}
Panoramic radiography is a widely used imaging modality in dental practice and research. However, it only provides flattened 2D images, which limits the detailed assessment of dental structures. In this paper, we propose Occudent, a framework for 3D teeth reconstruction from panoramic radiographs using neural implicit functions, which, to the best of our knowledge, is the first work to do so. For a given point in 3D space, the implicit function estimates whether the point is occupied by a tooth, and thus implicitly determines the boundaries of 3D tooth shapes. Firstly, Occudent applies multi-label segmentation to the input panoramic radiograph. Next, tooth shape embeddings as well as tooth class embeddings are generated from the segmentation outputs, which are fed to the reconstruction network. A novel module called Conditional eXcitation (CX) is proposed in order to effectively incorporate the combined shape and class embeddings into the implicit function. The performance of Occudent is evaluated using both quantitative and qualitative measures. Importantly, Occudent is trained and validated with actual panoramic radiographs as input, distinct from recent works which used synthesized images. Experiments demonstrate the superiority of Occudent over state-of-the-art methods.

\keywords{Panoramic radiographs  \and 3D reconstruction \and Teeth segmentation \and Neural implicit function}
\end{abstract}

\input{introduction.tex}
\input{methods.tex}
\input{experiments.tex}
\input{conclusion}

\textbf{Acknowledgements.}
This work was supported by the Korea Medical Device Development Fund grant funded by the Korea Government (the Ministry of Science and ICT, the Ministry of Trade, Industry and Energy, the Ministry of Health \& Welfare, the Ministry of Food and Drug Safety) (Project Number: 1711195279, RS-2021-KD000009); the National Research Foundation of Korea (NRF) Grant through the Ministry of Science and ICT (MSIT), Korea Government, under Grant 2022R1A5A1027646; the National Research Foundation of Korea (NRF) grant funded by the Korea government (MSIT) (No. 2021R1A2C1007215); the MSIT, Korea, under the ICT Creative Consilience program (IITP-2023-2020-0-01819) supervised by the IITP (Institute for Information \& communications Technology Planning \& Evaluation)

%
%
%
\bibliographystyle{splncs04}
\bibliography{reference}
\clearpage
\input{supplementary}

\end{document}

%% file: introduction.tex
\section{Introduction}
Panoramic radiography (panoramic X-ray, or PX) is a commonly used technique for dental examination and diagnosis. While PX produces 2D images from panoramic scanning, Cone-Beam Computed Tomography (CBCT) is an alternative imaging modality which provides 3D information on dental, oral, and maxillofacial structures. Despite providing more comprehensive information than PX, CBCT is more expensive and exposes patients to a greater dose of radiation \cite{brooks2009cbct}. Thus, 3D teeth reconstruction from PX is of significant value, e.g., 3D visualization can aid clinicians with dental diagnosis and treatment planning. Other applications include treatment simulation and interactive virtual reality for dental education \cite{li2021current}.

Previous 3D teeth reconstruction methods from 2D PX have relied on additional information such as tooth landmarks or tooth crown photographs. For example, \cite{mazzotta20132d} developed a model which uses landmarks on PX images to estimate 3D parametric models for tooth shapes, while \cite{abdelrehim20142d} reconstructed a single tooth using a shape prior and reflectance model based on the corresponding crown photograph.
Recent advances in deep neural networks have significantly impacted research on 3D teeth reconstruction. X2Teeth \cite{x2teeth} performs 3D reconstruction of the entire set of teeth from PX based on 2D segmentation using convolutional neural networks. Oral-3D \cite{song2021oral} generated 3D oral structures without supervised segmentation from PX using a GAN model \cite{goodfellow2020generative}. Yet, those methods relied on synthesized images as input instead of real-world PX images, where the synthesized images are obtained from 2D projections of CBCT \cite{arch}. The 2D segmentation of teeth from PX is useful for 3D reconstruction in order to identify and isolate teeth individually. Prior studies on 2D teeth segmentation \cite{koch2019accurate,zhao2020tsasnet} focused on binary segmentation determining the presence of teeth. However, this information alone is insufficient for the construction of individual teeth. Instead, we leverage recent frameworks \cite{nader2022automatic,silva2020study} on multi-label segmentation of PX into 32 classes including wisdom teeth.

In this paper, we propose \emph{Occudent}, an end-to-end model to reconstruct 3D teeth from 2D PX images. Occudent consists of a multi-label 2D segmentation followed by 3D teeth reconstruction using \emph{neural implicit functions} \cite{onet}. The function aims to learn the \textit{\textbf{occu}}pancy of \textit{\textbf{dent}}al structures, i.e., whether a point in space lies within the boundaries of 3D tooth shapes. Learning implicit functions is computationally advantageous over conventional encoder-decoder models outputting explicit 3D representations such as voxels, e.g., implicit models do not require large memory footprints to store and process voxels. Considering that 3D tooth shapes are characterized by tooth classes, we generate embeddings for tooth classes as well as segmented 2D tooth shapes. The combined class and shape embeddings are infused into the reconstruction network by a novel module called Conditional eXcitation (CX). CX performs learnable scaling of occupancy features conditioned on tooth class and shape embeddings. The performance of Occudent is evaluated with actual PX as input images, which differs from recent works using synthesized PX images \cite{song2021oral,x2teeth}. Experiments show Occudent outperforms state-of-the-art baselines both quantitatively and qualitatively. The main contributions are summarized as follows: (1) the first use of a neural implicit function for 3D teeth reconstruction, (2) novel strategies to inject tooth class and shape information into implicit functions, (3) the superiority over existing baselines which is demonstrated with real-world PX images.

%% file: methods.tex
\section{Methods}
\begin{figure} [t!]
\includegraphics[width=\textwidth]{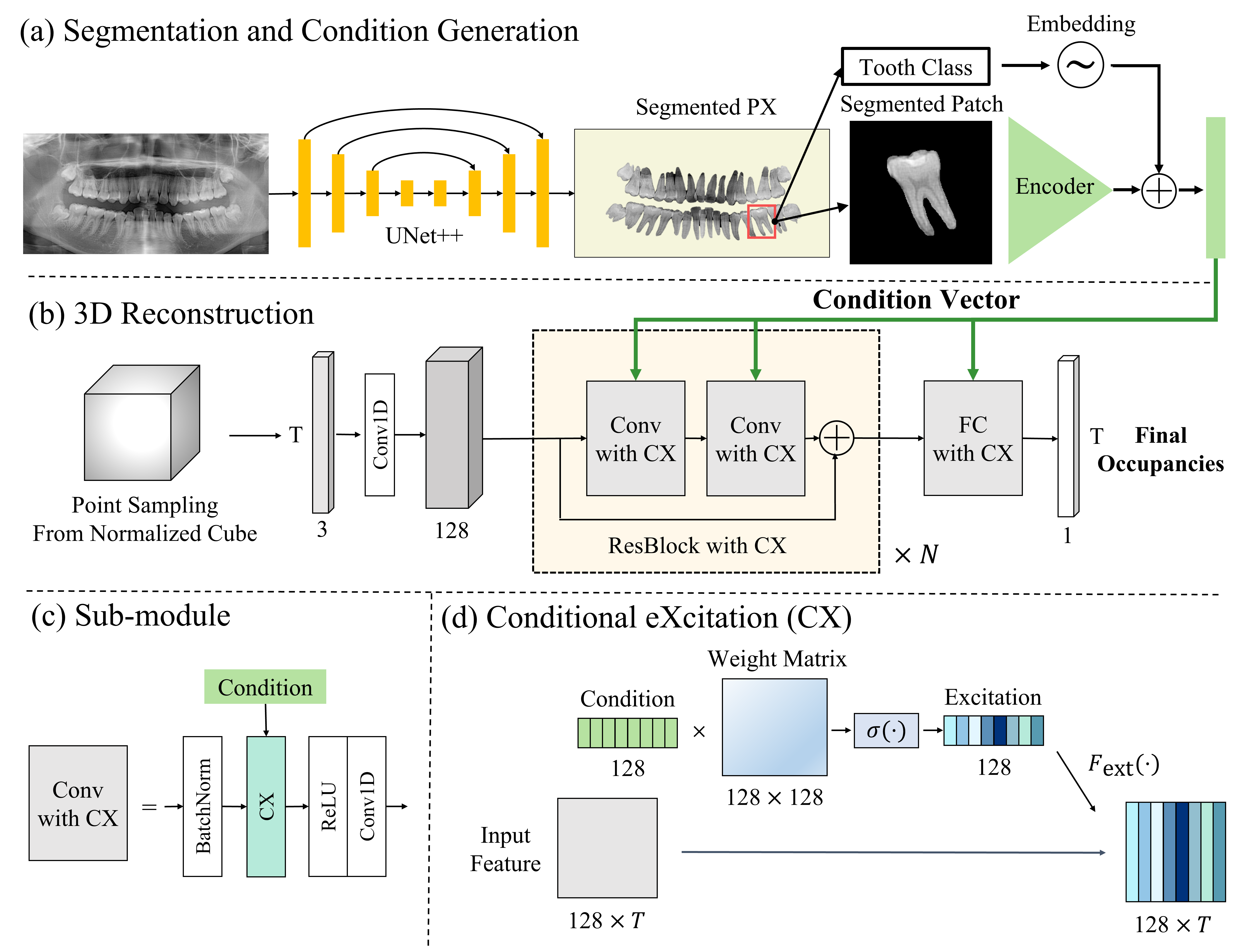}
\caption{(a) The PX image is segmented into 32 teeth classes using UNet\texttt{++} model. For each tooth, a segmented patch is generated by cropping input PX with the predicted segmentation mask, which is subsequently encoded via an image encoder. The tooth class is encoded by an embedding layer. The patch and class embeddings are added together to produce the condition vector. (b) The reconstruction process consists of $N$ ResBlocks to compute occupancy features of points sampled from 3D space. The condition vector from PX is processed by a Conditional eXcitation (CX) module incorporated in the ResBlocks. (c) The Conv with CX sub-module is composed of batch normalization, CX, ReLU, and a convolutional layer. The FC with CX is similar to the Conv with CX where Conv1D layer is replaced by fully connected layer. (d) CX injects condition information into the reconstruction network using excitation values. CX uses a trainable weight matrix to encode the condition vector into an excitation vector via a gating function. The input feature is scaled using the excitation vector through component-wise multiplication.} \label{model}
\end{figure}

The proposed model, Occudent, consists of two main components: 2D teeth segmentation and 3D teeth reconstruction. The former performs the segmentation of 32 teeth from PX using UNet\texttt{++} model \cite{zhou2018unetplusplus}. The individually segmented tooth and the tooth class are subsequently passed to the latter for the reconstruction. The reconstruction process estimates the 3D representation of the tooth based on a neural implicit function. The overall architecture of Occudent is depicted in Fig.~\ref{model}.

\subsection{2D Teeth Segmentation}
The teeth in input PX are segmented into 32 teeth classes. The 32 classes correspond to the traditional numbering of teeth, which includes incisors, canines, premolars and molars in both upper and lower jaws. We pose 2D teeth segmentation as a \emph{multi-label segmentation} problem \cite{nader2022automatic}, since nearby teeth can overlap with each other in the PX image, i.e., a single pixel of the input image can be classified into two or more classes.

The input of the model is $H \times W$ size PX image. The segmentation output has dimension $C \times H \times W$, where channel dimension $C=33$ represents the number of tooth classes: one class for the background and 32 classes for teeth similar to \cite{nader2022automatic}. Hence, the $H \times W$ output at each channel is a segmentation output for each tooth class. The segmentation outputs are used to generate tooth patches for reconstruction, which is explained later in detail. For the segmentation, we adopt pre-trained UNet\texttt{++} \cite{zhou2018unetplusplus} as the base model. UNet\texttt{++} is advantageous for medical image segmentation due to its modified skip pathways, which results in better performance compared to the vanilla UNet \cite{unet}.

\subsection{3D Teeth Reconstruction}
\subsubsection{Neural Implicit Representation.}
Typical representations of 3D shapes are point-based \cite{pointnet,psgn}, voxel-based \cite{r2n2,pix2vox}, or mesh-based methods \cite{pix2mesh}. These methods represent 3D shapes explicitly through a set of discrete points, vertices, and faces. Recently, implicit representation methods based on a continuous function which defines the boundary of 3D shapes have become increasingly popular \cite{nerf,onet,deepsdf}. Occupancy Networks \cite{onet} is a pioneering work which utilizes neural networks to approximate the implicit function of an object's occupancy. The term occupancy refers to whether a point in space lies in the interior or exterior of object boundaries. The occupancy function maps a 3D point to either 0 or 1, indicating the occupancy of the point. Let $o_\textrm{A}$ denote the occupancy function for an object $\textrm{A}$ as follows:
\begin{equation}
o_\textrm{A}: \rm I\!R^{3} \rightarrow \{0, 1\}
\end{equation}

In practice, $o_\textrm{A}$ can be estimated only by a set of observations of object $\textrm{A}$, denoted by $\mathcal{X}_\textrm{A}$. Examples of observations are projected images or point cloud data obtained from the object. Our objective is to estimate the occupancy function conditioned on $\mathcal{X}_\textrm{A}$. Specifically, we would like to find function $f_{\theta}$ which estimates the occupancy probability of a point in 3D space based on $\mathcal{X}_\textrm{A}$ \cite{onet}:
\begin{equation}\label{ftheta}
f_{\theta}:\rm I\!R^{3} \times \mathcal{X}_\textrm{A} \rightarrow [0, 1]
\end{equation}
Inspired by the aforementioned framework, we leverage segmented tooth patch and tooth class as observations denoted by condition vector $\bm{c}$. Specifically, the input to the function is a set of $T$ randomly sampled locations within a unit cube, and the function outputs the occupancy probability of the input. Thus, the function is given by $f_{\theta}:(x, y, z, \bm{c})\rightarrow [0, 1]$.

The model for $f_{\theta}$ is depicted in Fig.~\ref{model} (b). The sampled locations are projected to 128 dimensional feature vectors using 1D convolution. Next, the features are processed by a sequence of ResNet blocks followed by FC (fully connected) layers. Conditional vector $\bm{c}$ is used for each block through Conditional eXcitation (CX) which we will explain later. 

\subsubsection{Class-specific Conditional Features.}
A distinctive feature of the tooth reconstruction task is that teeth with the same number share properties such as surface and root shapes. Hence, we propose to use tooth class information in combination with a segmented tooth patch from PX. The tooth class is processed by a learnable embedding layer which outputs a class embedding vector. 

Next, we create a square patch of the tooth using the segmentation output as follows. A binary mask of the segmented tooth is generated by applying thresholding to the segmentation output. A tooth patch is created by cropping out the tooth region from the input PX, i.e., the binary mask is applied (bitwise AND) to the input PX to obtain the patch.
The segmented tooth patch is subsequently encoded using a pre-trained ResNet18 model \cite{he2016deep}, which outputs a patch embedding vector. The patch and class embeddings are added to yield the condition vector for the reconstruction model. This process is depicted in Fig.~\ref{model} (a).

Our approach differs from previous approaches, such as Occupancy Networks \cite{onet} which uses only single-view images for 3D reconstruction. X2Teeth \cite{x2teeth} also addresses the task of 3D teeth reconstruction from 2D PX. However, X2Teeth only uses segmented image features for the reconstruction. By contrast, Occudent leverages a class-specific encoding method to boost the reconstruction performance, as demonstrated in ablation analysis in Supplementary Materials.

\subsubsection{Conditional eXcitation.}
To effectively inject 2D observations into the reconstruction network, we propose Conditional eXcitation (CX) inspired by Squeeze-and-Excitation Network (SENet) \cite{hu2018squeeze}. In SENet, excitation refers to scaling input features according to their importance. In Occudent, the concept of excitation is extended to incorporating conditional features into the network. Firstly, the condition vector is encoded into excitation vector $\bm{e}$. Next, the excitation is applied to input feature by scaling the feature components by $\bm{e}$. The CX procedure can be expressed as:
\begin{equation}
\bm{e}=\alpha \cdot \sigma(W\bm{c}),
\end{equation}
\begin{equation}
\bm{y} = F_{\textrm{ext}}(\bm{e}, \bm{x})
\end{equation}
where $\bm{c}$ is the condition vector, $\sigma$ is a gating function, $W$ is a learnable weight matrix, $\alpha$ is a hyperparameter for the excitation result, and $F_{\textrm{ext}}$ is the excitation function. We use sigmoid function for $\sigma$, and component-wise multiplication for the excitation, $F_{\textrm{ext}}(\bm{e}, \bm{x}) = \bm{e}$ $\smallotimes$ $\bm{x}$. The CX module is depicted in Fig.~\ref{model} (d). CX differs from SENet in that CX derives the excitation from the condition vector, whereas SENet derives it from input features. Our approach also differs from Occupancy Networks which used Conditional Batch Normalization (CBN) \cite{dumoulin2016adversarially,de2017modulating} which combines conditioning with batch normalization. However, the conditioning process should be independent of input batches because those components serve different purposes in deep learning models. Thus, we propose to separate conditioning from batch normalization, as is done by CX.

%% file: experiments.tex
\section{Experiments}

\subsubsection{Dataset.}
The pre-training of the segmentation model was done with a dataset of 4000 PX images, sourced from \lq The Open AI Dataset Project (AI-Hub, S. Korea)\rq. All data information can be accessed through \lq AI-Hub (www.aihub.or.kr)\rq. The dataset consisted of two image sizes, 1976 $\times$ 976 and 2988 $\times$ 1468, which were resized to 256 $\times$ 768 to train the UNet\texttt{++} model.

For the main experiments for reconstruction, we used a set of 39 PX images and matched CBCT images, obtained from Korea University Anam Hospital. This study was approved by the Institutional Review Board at Korea University (IRB number: 2020AN0410). The panoramic radiographs were of dimensions 1536 $\times$ 2860 and were resized to 600 $\times$ 1200 and randomly cropped of 592 $\times$ 1184 size for the segmentation training. The CBCT images were of size 768 $\times$ 768 $\times$ 576, capturing cranial bones. The teeth labels for 2D PX and CBCT were manually annotated by two experienced annotators and subsequently verified by a board-certified dentist. To train and evaluate the model, the dataset was partitioned into training (30 cases), validation (2 cases), and testing (7 cases) subsets.

\subsubsection{Implementation Details.}
For the pre-training of the segmentation model, we utilized a combination of cross-entropy and dice loss. For the main segmentation training, we used only dice loss. The segmentation and reconstruction models were trained separately. Following the completion of the segmentation model training, we fixed this model to predict its output for the reconstruction model. 

Each 3D tooth label was fit in 144 $\times$ 80 $\times$ 80 size tensor which was then regarded as $[-0.5,0.5]^3$ normalized cube in 3D space. For the training of the neural implicit function, a set of $T=100,000$ points was sampled from the unit cube. The preprocessing was consistent with that used in \cite{Stutz2018ARXIV}. We trained all the other baseline models with these normalized cubes. For example, for 3D-R2N2 \cite{r2n2}, we voxelized the cube to $128^{3}$ size. For a fair comparison, the final meshes produced by each model were extracted and compared using four different metrics. The detailed configuration of our model is provided in Supplementary Materials.

\subsubsection{Baselines.} We considered several state-of-the-art models as baselines, including 3D-R2N2 \cite{r2n2}, DeepRetrieval \cite{retrieval,x2teeth}, Pix2Vox \cite{pix2vox}, PSGN \cite{psgn}, Occupancy Networks (OccNet) \cite{onet}, and X2Teeth \cite{x2teeth}. To adapt the 3D-R2N2 model to single-view reconstruction, we removed its LSTM component, following the approach in \cite{onet}. As for the DeepRetrieval method, we employed the same encoder architecture as 3D-R2N2, and utilized the encoded feature vector of the test image query. Subsequently, we compared each encoded vector from the test image to the encoded vectors from the training set, and retrieved the tooth with the minimum Euclidean distance of encoded vectors from the training set.

\begin{table}[t!]
\caption{Comparison with baseline methods. The format of results is \emph{mean}$\pm$\emph{std} obtained from 10 repetitions of experiments.}\label{tab1}
\centering
\setlength\tabcolsep{4.5pt}
\begin{tabular}{l|cccc}
\hline 
Method & IoU & Chamfer-$L_{1}$ & NC & Precision\\
\hline
3D-R2N2 & 0.585$\pm$0.005 & 0.382$\pm$0.008 & 0.617$\pm$0.009 & 0.634$\pm$0.010\\
PSGN & 0.606$\pm$0.015 & 0.342$\pm$0.016 & 0.829$\pm$0.012 & 0.737$\pm$0.018\\
Pix2Vox & 0.562$\pm$0.005 & 0.388$\pm$0.008 & 0.599$\pm$0.007 & 0.664$\pm$0.009\\
DeepRetrieval & 0.564$\pm$0.005 & 0.394$\pm$0.006 & 0.824$\pm$0.003 & 0.696$\pm$0.005\\
X2Teeth & 0.592$\pm$0.006 & 0.361$\pm$0.017 & 0.618$\pm$0.002 & 0.670$\pm$0.009\\
OccNet & 0.611$\pm$0.006 & 0.353$\pm$0.008 & 0.872$\pm$0.003 & 0.691$\pm$0.011\\
\textbf{Occudent (Ours)} & \textbf{0.651}$\pm$\textbf{0.004} & \textbf{0.298}$\pm$\textbf{0.006} & \textbf{0.890}$\pm$\textbf{0.001} & \textbf{0.739}$\pm$\textbf{0.008}\\
\hline
\end{tabular}
\end{table}

\subsubsection{Evaluation Metrics.}
The evaluation of the proposed method was conducted using the following metrics: volumetric Intersection over Union (IoU), Chamfer-$L_{1}$ distance, and Normal Consistency (NC), as outlined in prior work \cite{onet}. In addition, we used volumetric precision \cite{x2teeth} as a metric given by $| D \cap G|/|D|$
where $G$ denotes the ground-truth set of points occupied by the object, and $D$ denotes the set of points predicted as the object.

\subsubsection{Quantitative Comparison.}
Table \ref{tab1} presents a quantitative comparison of the proposed model with several baseline models. The results demonstrate that Occudent surpasses the other methods across all the metrics, and the methods based on neural implicit functions (Occudent and OccNet) perform better compared to conventional encoder-decoder approaches, such as Pix2Vox and 3D-R2N2.
The performance gap between Occudent and X2Teeth is presumably because real PX images are used as input data. X2Teeth used \emph{synthesized} images generated from the 2D projections of CBCT in \cite{arch}. Thus, both the input 2D shape and the target 3D shape come from the same modality (CBCT). However, the distribution of real PX images may differ significantly from that of 2D-projected CBCT. Explicit methods can be more sensitive to such differences than implicit methods, because typically in explicit methods, input features are directly encoded and subsequently decoded to predict the target shapes \cite{r2n2,pix2vox,x2teeth}. Overall, the differences in the IoU performances among the baselines are somewhat small. This is because all the baselines are moderately successful in generating coarse tooth shapes. However, Occudent is significantly better at generating details such as root shapes, which will be shown in the subsequent section.

\begin{figure}[t!]
\includegraphics[width=\textwidth]{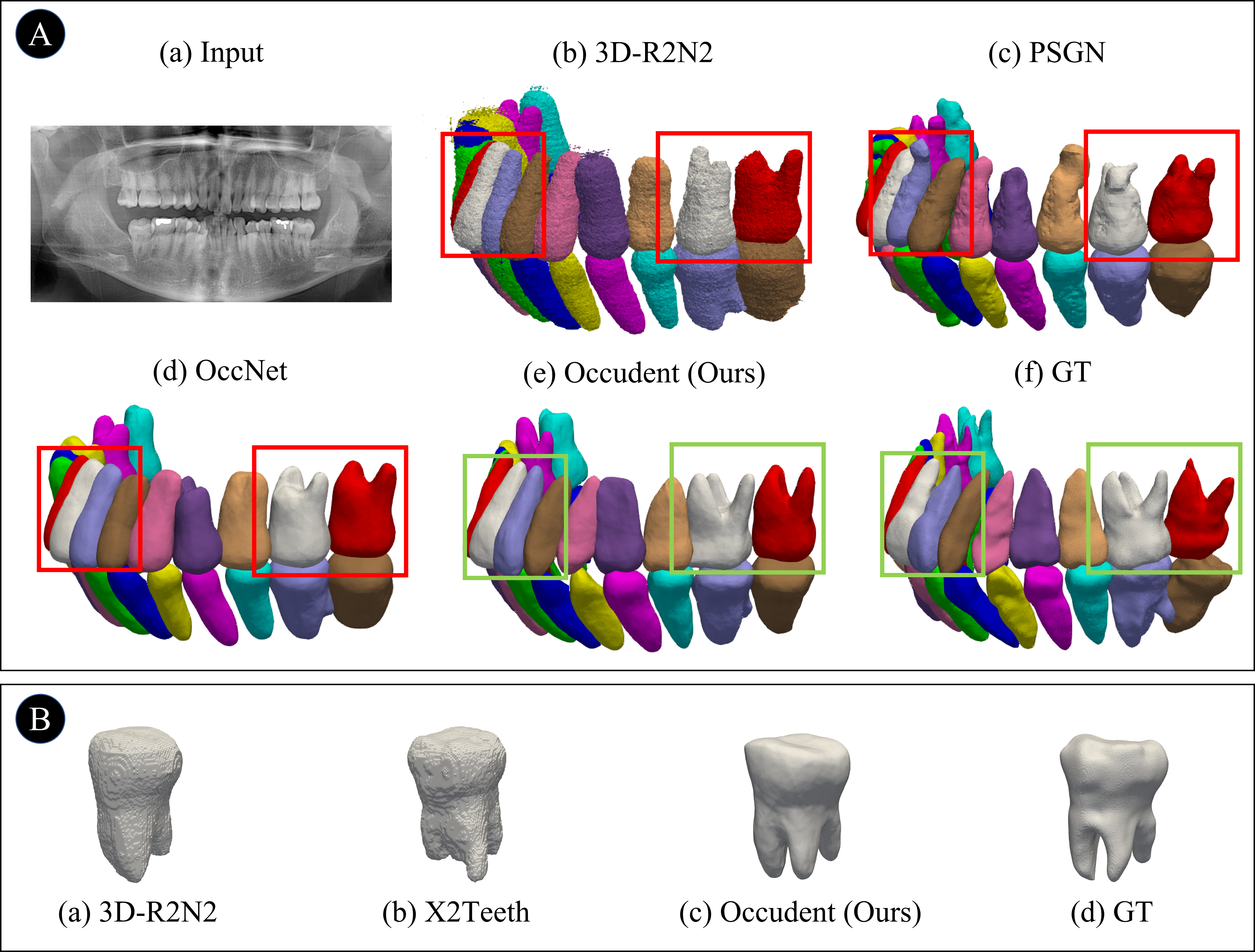}
\caption{Visual representation of sample outputs. Boxes are used to highlight the incisors and molars in the upper jaw.} \label{visual}
\end{figure}

\subsubsection{Qualitative Comparison.}
Fig. \ref{visual}A illustrates the qualitative results of the proposed method in generating 3D teeth mesh outputs. From our model, each tooth is generated, and generated teeth are combined along with an arch curve based on a beta function \cite{braun1998form}. Fig. \ref{visual}A demonstrates that our proposed method generates the most similar-looking outputs compared to the ground truth. For instance, our model can reconstruct a plausible shape for all tooth types including detailed shapes of molar roots. 3D-R2N2 produces larger and less detailed tooth shapes. PSGN and OccNet are better at generating rough shapes than 3D-R2N2, however, lack in detailed root shapes.

As illustrated in Fig. \ref{visual}B, Occudent produces a more refined mesh of tooth shape representation than voxel-based methods like 3D-R2N2 or X2Teeth. One of the limitations of voxel-based methods is that they heavily depend on the resolution of the output. For example, increasing the output size leads to an exponential increase in model size. By contrast, Occudent employs continuous neural implicit functions to represent shape boundaries, which enables us to generate smoother output and to be robust to the target size.

%% file: conclusion.tex
\section{Conclusion}
In this paper, we present a framework for 3D teeth reconstruction from a single PX. To the best of our knowledge, our method is the first to utilize a neural implicit function for 3D teeth reconstruction. The performance of our proposed framework is evaluated quantitatively and qualitatively, demonstrating its superiority over state-of-the-art techniques. Importantly, our framework is capable of accommodating two distinct modalities, PX, and CBCT. Our framework has the potential to be valuable in clinical practice and also can support virtual simulation or educational tools. In the future, further improvements can be made, such as incorporating additional imaging modalities or exploring neural architectures for more robust reconstruction.

%% file: supplementary.tex
\section*{Model Configuration}
\vspace{-0.7cm}
\begin{table}
\renewcommand\thetable{S1}
\caption{Detailed model configuration of Occudent.}
\centering
\setlength\tabcolsep{4.5pt}
\begin{tabular}{|l|c|}
\hline
\# of ResBlocks with CX & 5\\
\hline
Hidden feature size & 128\\
\hline
Condition vector dimension & 128\\
\hline
Optimizer & Adam optimizaer\\
\hline
Learning rate & 1e-4\\
\hline
Batch size & 10\\
\hline
Training epochs & 200 to 300\\
\hline
Usage of early stopping method & Yes\\
\hline
GPU for pre-training & NVIDIA Tesla V100\\
\hline
GPU for main experiments & NVIDIA A100 and NVIDIA RTX A6000\\
\hline
\end{tabular}
\end{table}

\section*{Ablation Study}

\vspace{-0.7cm}
\begin{table}
\renewcommand\thetable{S2}
\caption{\textbf{Ablation study on Conditional eXcitation (CX) and tooth class embeddings.} We demonstrate the effectiveness of two components of our model: CX and the inclusion of tooth class embeddings. We compare CX with the case where Conditional Batch Normalization (CBN) \cite{dumoulin2016adversarially,de2017modulating} is used instead, as in the Occupancy Network \cite{onet}. A comparison of the 1st and 2nd rows and the 3rd and 4th rows shows that CX outperforms CBN irrespective of using tooth class embeddings. By comparing the 1st and 3rd rows and the 2nd and 4th rows, we observe that incorporating tooth class embeddings into reconstruction improves the model's performance.}
\centering
\setlength\tabcolsep{4.5pt}
\begin{tabular}{l|cccc}
\hline
 & IoU & Chamfer-$L_{1}$ & NC & Precision\\
\hline
CBN Only& 0.611$\pm$0.006 & 0.353$\pm$0.008 & 0.872$\pm$0.003 & 0.691$\pm$0.011\\
CX Only& 0.621$\pm$0.003 & 0.336$\pm$0.005 & 0.875$\pm$0.001 & 0.701$\pm$0.006\\
CBN+Tooth Class & 0.641$\pm$0.010 & 0.311$\pm$0.013 & 0.886$\pm$0.005 & 0.728$\pm$0.011\\
\textbf{CX+Tooth Class} & \textbf{0.651}$\pm$\textbf{0.004} & \textbf{0.298}$\pm$\textbf{0.006} & \textbf{0.890}$\pm$\textbf{0.001} & \textbf{0.739}$\pm$\textbf{0.008}\\
\hline
\end{tabular}
   
\end{table}